\documentclass[letterpaper, 10 pt, conference]{ieeeconf}  

\IEEEoverridecommandlockouts                              

\usepackage{graphics} 
\usepackage{epsfig} 
\usepackage{mathptmx} 
\usepackage{times} 
\usepackage{amsmath} 
\usepackage{amssymb}  
\usepackage{multirow}
\usepackage{multicol}
\usepackage{booktabs}
\usepackage{hyperref}
\usepackage{cleveref}
\usepackage{cite}
\usepackage{censor}
\usepackage{svg}
\usepackage{subcaption}
\usepackage{colortbl}
\let\labelindent\relax
\usepackage[inline]{enumitem}
\DeclareMathAlphabet{\mathcal}{OMS}{cmsy}{m}{n}
\crefname{figure}{Fig.}{Figs.}
\Crefname{figure}{Fig.}{Figs.}

\title{\LARGE \bf
MAMMOTH: A Multi-Modal End-to-End Policy for Off-Road Mobility Robust to Missing Modality
}

\author{Ahaan Kotian$^{1}$, Shivani Subramanyan$^{1}$ and Suresh Sundaram$^{1}$
\thanks{$^{1}$Artificial Intelligence and Robotics Lab, Department of Aerospace Engineering, Indian Institute of Science, India. 
        {\tt\small ahaan.kotianstudent@gmail.com}}%
}

\begin{document}

\maketitle
\thispagestyle{empty}
\pagestyle{empty}

\begin{abstract}

Reliable autonomous navigation in unstructured off-road environments remains a critical unsolved challenge due to extreme terrain diversity, drastic illumination variations and acute sensor degradation. Recent developments have approached the problem as a traversability costmap estimation or visual navigation task. However, many exhibit heavy reliance on RGB modality, leading to poor performance in varied illumination such as glares, shadows or low ambient light. Achieving robust generalization in such conditions requires integrating modalities that provide supplementary scene information. Such multi-modal methods suffer from a rigid dependency on the presence of near-perfect sensor inputs, leaving them unable to robustly handle sensor degradation or individual modality failure. To address these limitations, we introduce MAMMOTH (MAsking Multi-Modal inputs for Off-road Traversability Heuristic-informed navigation), a unified end-to-end navigation policy for robust off-road visual-goal-conditioned navigation and undirected exploration. Specifically, MAMMOTH efficiently fuses multi-modal observations (RGB, Thermal, 3D Pointcloud and Ego Velocity) and is trained with a modality dropout scheme, enabling it to generalize to missing modalities at inference time. Furthermore, we employ a diffusion policy to learn the joint conditional probability distribution of physically-grounded trajectories and a intrinsic traversability heuristic. MAMMOTH utilizes this heuristic to prefer safer, smoother trajectories. We validate MAMMOTH through extensive real-world robot experiments in distinct off-road environments, including night-time operation. Our results demonstrate superior performance, with significant improvements in collision avoidance, terrain-aware planning and generalization to missing modalities. The code and dataset used for this work will be made publicly available.

\end{abstract}

\section{INTRODUCTION}

Off-road autonomous navigation presents various complex challenges that require pushing the limits of robotic perception and planning. These environments are often unstructured and characterized by diverse lighting conditions like glares and shadows, as well as extremely low ambient light found in subterranean caves or night-time environments. High uncertainty in observations makes information from any individual modality unreliable or insufficient depending on the scene, especially in visually degraded conditions. This is compounded by complex terrain and presence of unpredictable obstacles like ditches, sharp rocks, tree roots etc. Furthermore, these environments present a persistent risk of acute sensor degradation due to adverse weather conditions, high vibrations and low illumination. Classical navigation approaches have been proposed but are not equipped for such environments. They rely heavily on strong geometric priors and usually have a strict requirement of pre-built maps. This renders them unfit for scalability and generalization to novel environments.

Recently, learning-based methods have advanced the field, replacing manually crafted features with data-driven representations. Research in this area typically falls in two categories: traversability estimation approaches and end-to-end (E2E) policies. Traversability estimation approaches such as SALON \cite{sivaprakasam2025salon}, HDIF \cite{10160856}, RoadRunner M\&M \cite{10740919} and IRisPath \cite{sharma2025irispathenhancingcostmapoffroad} demonstrate the use of multi-modal fusion for estimating costmaps based on traversability. While these methods show great success, they suffer from two crucial problems: (i) these methods do not generalize to low-light conditions (ii) they possess no capability to deal with acute sensor degradation which is a prevalent phenomenon in off-road environments. Conversely, E2E policies like ViNT \cite{shah2023vint}, NoMaD \cite{sridhar2024nomad} etc. have a strong coupling between sensor inputs and actions, offering reactive policies. Many of these policies are trained on large-scale diverse datasets resulting in intelligent behaviors. However, these policies are typically designed using only RGB modality. This heavy reliance on visual inputs makes them unreliable for varied lighting conditions and visually degraded environments which are endemic in off-road environments. 
More recent approaches like DTG \cite{10802055}, MTG \cite{liang2024mtg} utilize pointclouds from LiDAR sensors, but are still vulnerable to missing modality scenarios. 
    
To address these limitations, we introduce MAMMOTH (MAsking Multi-Modal inputs for Off-road Traversability Heuristic-informed navigation), an E2E navigation policy capable of visual-goal-conditioned navigation and undirected exploration in off-road environments while being robust to drastic lighting variations and acute sensor degradation. MAMMOTH exhibits system-level robustness and high deployability by remaining operational in diverse settings which RGB-based E2E policies and costmap estimation methods do not solve for. MAMMOTH’s capabilities are a result of 3 core design principles: 

\begin{enumerate}
\item Synergistic multi-modal fusion: MAMMOTH takes input from a comprehensive suite of RGB images, thermal LWIR images, 3D LiDAR pointclouds and ego velocity information, fusing them to leverage the strengths of each modality and the complementary scene understanding they provide. MAMMOTH learns to benefit from both selective reliance (which combination of modalities to "weigh" more) and synergistic fusion (deep cross-modal understanding) of multiple modalities depending on the perceived scene by using a sparse Mixture-of-Experts (MoE) backbone \cite{shazeer2017outrageously}. This results in robust mobility in both day and night conditions.

\item Flexible and robust deployment: We train MAMMOTH using a random modality dropout strategy where we mask out all tokens of a modality or a combination of modalities, forcing the model to learn cross-modal representations and preventing it from over-relying on any specific modality. This approach directly translates to zero-shot generalization to real-world missing modality scenarios due to acute sensor degradation during deployment. Furthermore, this rewards us with system-level flexibility where MAMMOTH now has the capability to be seamlessly deployed on robotic platforms that may not possess the full sensor suite (e.g. lacking LWIR camera or LiDAR sensor) and give appropriate performance.

\item Physically-Aware Trajectory Generation: To navigate complex off-road environments, a heuristic about the terrain proves to be critical. While recent costmap estimation or guidance methods require higher computational budget, we task a diffusion policy to estimate the joint conditional probability distribution over physically-grounded trajectories and an associated intrinsic traversability heuristic derived from IMU signals. This allows MAMMOTH to prefer smoother, safer trajectories with negligible computational overhead, making it a more elegant and efficient solution than external, post-hoc guidance.
\end{enumerate}

MAMMOTH's primary contributions are system-level robustness and high deployability. To the best of our knowledge, MAMMOTH is the first E2E policy to demonstrate robust mobility in complete night-time or low ambient light off-road environments and high-fidelity generalization to missing modality scenarios. We conduct extensive experiments in distinct off-road environments, demonstrating that MAMMOTH outperforms existing baselines in both visual-goal-conditioned navigation and undirected exploration. We validate our contributions through a rigorous series of ablation studies, precisely quantifying the impact of modality dropout, the fusion technique, and the efficacy of our traversability heuristic.

\begin{figure*}[tbp]
    \centering 
    \includegraphics[width=0.9\textwidth]{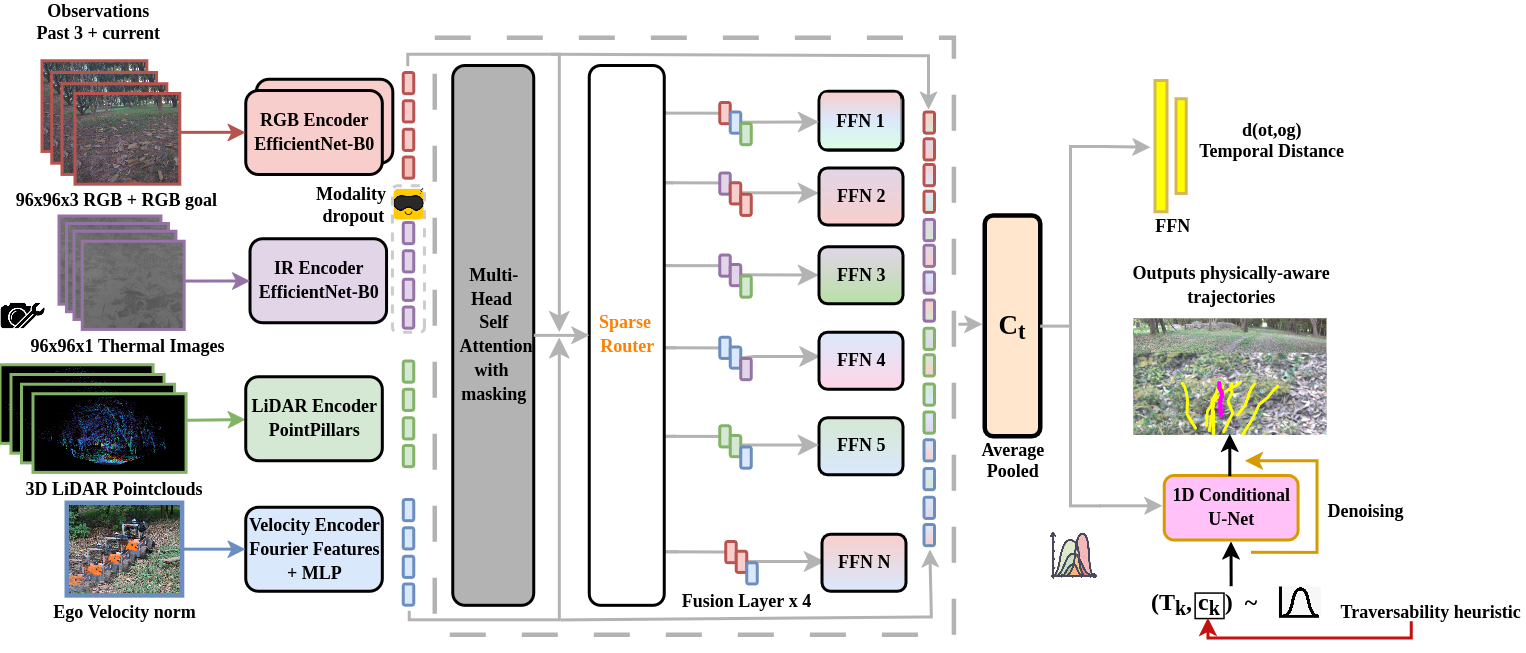} 
    \captionsetup{font=footnotesize}
    \caption{\textbf{Model architecture of MAMMOTH.} The architecture leverages distinct encoders for RGB, Thermal, 3D pointcloud, and ego velocity inputs. In the fusion layer, a sparse router dynamically selects and fuses features by feeding into expert networks. A diffusion policy then generates physically grounded multi-modal trajectories while a MLP predicts the temporal distance to the goal. }
    \vspace{-14pt}
    \label{fig:model_arch}
\end{figure*}

\section{Related Works}
\textbf{End-to-End Policies for Real-World Navigation.}
E2E learning \cite{chen2021interpretable, park2025endtoenddrivingselfsupervisedimitation, 10658504} has emerged as a promising direction in robot navigation, aiming to map raw sensory inputs directly to control outputs or local waypoints. 
Generalist policies such as GNM \cite{shah2023gnm} and ViNT \cite{shah2023vint} have shown that a single goal-conditioned model, trained on diverse multi-robot datasets, can generalize across environments and embodiments~\cite{yang2024pushinglimitscrossembodimentlearning, eftekharone}.
Methods like ViKiNG \cite{Shah_2022}, and RECON \cite{shah2022rapid} extend this idea to long-range goal-conditioned navigation. 
More recently, NoMaD \cite{sridhar2024nomad} introduces a unified diffusion-based policy for both visual goal-conditioned navigation and undirected exploration, while  FlowNav \cite{gode2025flownav} builds upon NoMaD \cite{sridhar2024nomad} by incorporating depth priors from visual foundation models as an additional modality. 
Language-conditioned navigation models such as OmniVLA \cite{hirose2025omnivlaomnimodalvisionlanguageactionmodel}, LM-Nav \cite{shah2023lm}, and LeLAN \cite{hiroselelan} further expand generalist navigation capabilities by integrating natural language as an auxiliary modality, achieving strong cross-domain generalization \cite{hirose2025learningdrivemodelbasedreannotation}.
Despite these advances, most of these approaches remain specialized to structured indoor or outdoor settings and often fail in unstructured off-road environments due to their lack of terrain-awareness and reliance on RGB inputs, which are highly sensitive to lighting and appearance variations. 
LiDAR pointcloud-based methods such as MTG \cite{liang2024mtg} and DTG \cite{10802055} improve long-range planning by leveraging pointclouds and ego-velocity cues, yet remain vulnerable to missing modality scenarios and cannot explicitly estimate terrain traversability - both critical for reliable off-road navigation. 
Motivated by these limitations, we design an E2E policy that fuses multiple sensing modalities and remains robust across varying input combinations and adverse terrain conditions.

\textbf{Learning-Based Traversability Estimation.}
Learning-based approaches have increasingly replaced hand-crafted traversability heuristics with deep networks that infer terrain cost directly from sensor data or learn the kinodynamic relationship of the terrain~\cite{datar2024terrain, nazeri2025verticoder}. 
SALON~\cite{sivaprakasam2025salon} learns risk-aware cost and speed maps from limited off-road data, while HDIF~\cite{10160856} fuses visual, pointcloud, and proprioceptive cues to estimate terrain difficulty. 
IRisPath~\cite{sharma2025irispathenhancingcostmapoffroad} integrates thermal imaging for illumination-invariant traversability prediction.
RoadRunner M\&M~\cite{10740919} utilizes camera–LiDAR fusion for long-range elevation and cost estimation while MOSU~\cite{liang2025mosuautonomouslongrangerobot} leverages RGB and 3D pointclouds from LiDAR for off-road navigation but relies on computationally expensive vision–language models for ranking trajectories and lacks robustness to missing modalities. 
While these methods highlight the importance of multi-modal inputs and traversability reasoning, they assume near-perfect conditions and degrade significantly under domain shifts or sensor dropout.

Generative models have recently been adopted for trajectory generation. 
NaviDiffusor~\cite{zeng2025navidiffusor}, Diffusion-ES~\cite{yang2024diffusion}, \cite{thakker2025risk} demonstrate that diffusion policies guided by geometric, cost-based or risk-aware priors can generate smooth, feasible trajectories. 
However, such guidance signals are often unavailable or computationally expensive in real-world off-road settings. 
In contrast, MAMMOTH introduces a diffusion-based policy that learns to generate cost-consistent trajectories directly from sensory observations. 

\section{Methodology}
\subsection{Formulation}
Our objective is to learn a unified control policy $\pi$ that maps a history of $k$ multi-modal observations from time $t$ $O_t = \{o_{t-k}, \dots, o_t\}$ to a distribution over future actions $\mathcal{A}$. A primary requirement is that $\pi$ must be robust to arbitrary modality dropouts, generalizing its inference to any available subset of the input sensor streams.
The action space $\mathcal{A}$ is a distribution over trajectories, where each trajectory $\mathcal{T} \in \mathcal{A}$ is a sequence of $N$ waypoints, $\mathcal{T} = \{ (x_i, y_i, c_i) \}_{i=1}^N$. Here, $(x_i, y_i)$ are the local 2D coordinates of the waypoint, and $c_i$ is our intrinsic traversability heuristic. This heuristic allows the policy to evaluate the physical consequences of its generated plans.
Following the setup of NoMaD \cite{sridhar2024nomad}, this policy operates in two modes. For visual-goal-conditioned navigation, the policy $\pi(O_t, o_g)$ is additionally conditioned on a goal image $o_g$ and must generate actions that safely guide the robot to the goal. For undirected exploration, $o_g$ is unavailable, and the policy $\pi(O_t)$ must generate safe and efficient exploration behaviors (e.g., obstacle avoidance, terrain-aware path selection). We pair this unified policy with a topological memory $\mathcal{M}$ \cite{Shah_2022} as a high-level planner to guide long-horizon navigation.

\subsection{MAMMOTH}
\textbf{Overview.}
MAMMOTH is a unified E2E navigation policy designed to fuse multi-modal sensory inputs efficiently, generate physically-grounded trajectories, and remain deployable when one or more sensing modalities are severely degraded.
As illustrated in Figure~\ref{fig:model_arch}, the model consists of modality-specific encoders, fusion layers for cross-modal fusion, and a diffusion-based trajectory head. 
We next describe its design decisions in detail, starting with our multi-modal fusion mechanism.

{\bf Synergistic Multi-Modal Fusion.}
In order to achieve robust navigation capability in off-road environments, we fuse multiple input modalities from a diverse sensor suite consisting of RGB images, thermal images, 3D LiDAR pointclouds and ego velocity information. MAMMOTH fuses these modalities to leverage the complementary strengths that these individual modalities provide. MAMMOTH takes as input the current and past $k$ observations $O_t = \{o_{t-k}, \dots, o_t\}$ from each of these $M$ modalities and encodes each observation into a single latent representation $z \in \mathbb{R}^D$. These latent representations are then fused in the fusion layers where each layer consists of a Multi-Head Self Attention (MHSA) 
block coupled with a sparse MoE layer. We utilize the per-modality router design with Laplace gating from \cite{han2024fusemoe}. This setup uses four lightweight routers (one for each modality) to dynamically select a $topk$ subset from a shared pool of $N$ expert networks.
This conditional computation architecture with the unique router design and gating is key for two reasons. First, it enables experts to effectively fuse information from different modalities while balancing specialization of experts to specific modalities as shown in \cite{han2024fusemoe}. Consequently, MAMMOTH builds cross-modal understanding through synergistic fusion, as well as the capability to exhibit selective reliance by "weighing" observations from certain modalities more than others depending on the perceived environmental conditions. Second, it speeds up inference and reduces memory bandwidth by decoupling the model's total parameters from its per-inference FLOPs.

{\bf Flexible and Robust Deployment.}
While the sparse MoE backbone handles multi-modal fusion, it does not inherently solve for acute sensor degradation which is a frequent occurrence in complex off-road environments. This can be caused by high vibrations due to undulated, rough terrain, high lighting variation, or adverse weather conditions. To enable MAMMOTH to robustly handle missing modalities arising from such scenarios, we incorporate modality dropout \cite{nezakati2025mmp} during training. During the training phase, we sample a binary mask $m \in \{0, 1\}$ for each modality from a Bernoulli distribution. A mask of $m=1$ signifies that all tokens corresponding to that modality are dropped for that specific training step. During inference, we can deterministically set these masks to $0$ or $1$ to simulate any combination of sensor failures. In contrast to MMP \cite{nezakati2025mmp}, we do not attempt to reconstruct the masked tokens, thereby avoiding the significant computational overhead of the reconstruction objective. Our "mask-and-ignore" approach achieves the desired robustness while remaining lightweight enough for real-time inference on our physical robot. The combination of the unique sparse MoE backbone and this masking strategy allows MAMMOTH to generalize effectively to missing modalities while retaining efficiency. Additionally, this flexibility is a key practical advantage, as it allows our model to be seamlessly deployed on robotic platforms that may not possess the full sensor suite (e.g., lacking LWIR camera or LiDAR sensor).

{\bf Physically-Aware Trajectory Generation.}
A key contribution of our work lies in moving beyond simple geometric planning. For complex off-road environments, a viable plan must account for the physical consequences of traversing the terrain, such as vehicle stability. We achieve this by tasking a diffusion policy with learning the joint conditional probability distribution $p(\mathcal{T} | \mathbf{C}_t)$ over \textit{physically-grounded} trajectories $\mathcal{T}$, conditioned on our rich multi-modal sensory and goal context $\mathbf{C}_t$. Crucially, our trajectory representation $\mathcal{T}$ is augmented with a \textit{self-supervised traversability heuristic}, which we derive from the Power Spectral Density (PSD) of the vertical (Z-axis) acceleration from onboard IMU data~\cite{sharma2025irispathenhancingcostmapoffroad}. By formulating this as a conditional Denoising Diffusion Probabilistic Model (DDPM) \cite{ho2020denoising}, our model is trained to reverse a $T$-step noising process. It learns to predict the injected noise $\boldsymbol{\epsilon}$ from a noisy sample $\mathbf{x}_t = \sqrt{\bar{\alpha}_t} \mathbf{x}_0 + \sqrt{1 - \bar{\alpha}_t} \boldsymbol{\epsilon}$ by optimizing a network $\boldsymbol{\epsilon}_\theta(\mathbf{x}_t, t, \mathbf{C}_t)$ using the loss
\begin{align}
\mathcal{L}_{\text{Diff}} = \mathbb{E}_{t, \mathbf{x}_0, \boldsymbol{\epsilon}} \left[ || \boldsymbol{\epsilon} - \boldsymbol{\epsilon}_\theta(\mathbf{x}_t, t, \mathbf{C}_t) ||^2 \right].
\end{align}
This conditioning is vital, as it forces the model to learn the intricate, non-linear correlation between a spatial path and its associated physical cost. This allows for superior trajectory evaluation at inference, where we use a \textit{trajectory ranking} method that ranks $N$ candidate trajectories using the computed traversability heuristic of each trajectory to prioritize vehicle safety and stability.

\subsection{Loss function}
MAMMOTH is trained end-to-end in a supervised fashion with the loss function
\vspace{-3pt}
\begin{align}
\mathcal{L} = \alpha \cdot \mathcal{L}_{\text{Dist}} + (1 - \alpha) \cdot \beta \cdot \mathcal{L}_{\text{Diff}} + (1 - \beta) \cdot \mathcal{L}_{\text{MoE}},
\end{align}
where $\mathcal{L}_{\text{Dist}}$ is the L2 loss of the temporal distance prediction and $\alpha$ and $\beta$ are hyperpameters to define the trade-off between the terms of the loss function. $\mathcal{L}_{\text{MoE}}$ is the auxillary load-balancing loss which encourages balanced utilization of all experts. It is defined as:
\vspace{-3pt}
\begin{align}
\mathcal{L}_{\text{MoE}} = \lambda_{\text{aux}} \cdot \left( \text{CV}^2(I) + \text{CV}^2(L) \right),
\end{align}
where $\mathcal{\lambda_{\text{aux}}}$ is a hyperparamter that scales the loss while $I$ and $L$ denote expert importances and expert loads respectively.

\begin{figure*}[ht]
    \centering 
    
    \begin{subfigure}[b]{0.3\textwidth}
        \centering
        \includegraphics[width=\linewidth]{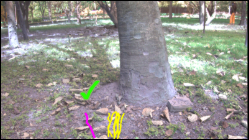}
        \captionsetup{font=footnotesize}
        \caption{MAMMOTH}
        \label{fig:model_arch_sub}
    \end{subfigure}
    \hfill 
    \begin{subfigure}[b]{0.3\textwidth}
        \centering
        \includegraphics[width=\linewidth]{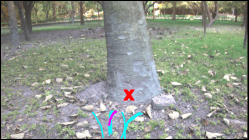} 
        \captionsetup{font=footnotesize}
        \caption{NoMaD} 
        \label{fig:other_image_sub}
    \end{subfigure}
    \hfill 
    \begin{subfigure}[b]{0.3\textwidth}
        \centering
        \includegraphics[width=\linewidth]{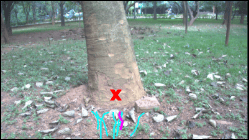}
        \captionsetup{font=footnotesize}
        \caption{FlowNav} 
        \label{fig:third_image_sub}
\end{subfigure}

    \captionsetup{font=footnotesize}
        \caption{\textbf{Results of baseline comparison.} Comparison of MAMMOTH with NoMaD and FlowNav. Pink indicates the chosen trajectory, while yellow/blue shows candidate trajectories. (a) MAMMOTH avoids the obstacle while (b) NoMaD and (c) FlowNav collide with the obstacle. }
        \vspace{-16pt}
    \label{fig:comparison} 
\end{figure*}

\section{Experiments}
{\bf Datasets.}
We train MAMMOTH and all baselines on a combined dataset comprising the public TartanDrive 2.0 \cite{sivaprakasam2024tartandrive} 
and a new in-house off-road dataset, which we plan to release soon. This combined corpus provides over 9 hours of off-road navigation data. Although, TartanDrive 2.0 offers diverse off-road environments, it notably lacks night-time data and a thermal LWIR modality. Our in-house dataset was collected to fill this gap, adding the thermal LWIR modality and providing trajectories in challenging low-light and night-time conditions. 
For sensor inputs, we use the forward-facing Livox LiDARs and frontal RGB cameras from both datasets. The thermal LWIR modality is sourced exclusively from our in-house data.
We generate ground truth trajectories from the odometry data in both datasets, computed relative to the robot's local frame. Our traversability heuristic label is consistently computed for all trajectories using the Z-axis IMU signal.

{\bf Implementation Details.}
We process RGB and thermal inputs as $96 \times 96$ images, which are tokenized using separate EfficientNet-B0 
encoders. The norm of the ego velocity vector is encoded using Fourier features from \cite{sharma2025irispathenhancingcostmapoffroad}, which increases the input velocity’s dimensionality and enhances high-frequency patterns. These velocity embeddings are then processed using a MLP to produce a vector representation.
For the 3D pointcloud modality, we process single scans of 48,000 points (4-channel: $x, y, z, \text{intensity}$) using a PointPillars \cite{lang2019pointpillars} encoder. The encoder is configured with a voxel size of ($V_x, V_y, V_z$) = ($0.32, 0.32, 4.0$) m and a point cloud range of [$0, 25.6$] m (X-axis) and [$-17.92, 17.92$] m (Y-axis). This produces a BEV feature map that is subsequently distilled into a vector representation.
For all modalities, we use a context of 4 observations ($k=3$ past, 1 current), which are tokenized into 256-dimensional embeddings using modality-specific encoders. This is processed with 4 fusion layers and 4 attention heads for the MHSA block per layer. Each layer has $N=16$ experts and $M=4$ routers (one each for RGB, Thermal, 3D pointcloud and Velocity modalities) that routes tokens to this shared pool of experts. The diffusion policy generates 8 trajectories with a temporal horizon of 8 waypoints using a conditional Denoising Diffusion Probabilistic Model (DDPM) \cite{ho2020denoising} with a cosine noise scheduler and $T=10$ denoising timesteps. The noise prediction network is a 1-D Conditional U-Net \cite{chi2023diffusionpolicy}.

During training, the RGB, 3D pointcloud and Velocity modalities are masked with a probability of $p=0.5$, distributed in equal proportion across all masking combinations. The thermal modality uses a lower masking probability of $p=0.2$. This is a deliberate choice to compensate for the lack of thermal modality in TartanDrive 2.0, ensuring the model learns to effectively utilize the Thermal stream when it is present and not masked.
We train MAMMOTH with an initial learning rate of $10^{-4}$ for 20 epochs with a batch size of 100 on a NVIDIA H100 GPU. Each dataset is split into $80\%/20\%$ for train/test. We use the AdamW optimizer with cosine scheduling and warmup for stable training.

\begin{figure*}[t]
    \centering
    
    \begin{minipage}[c]{0.35\textwidth}
        \centering
        \includegraphics[height=4.6cm]{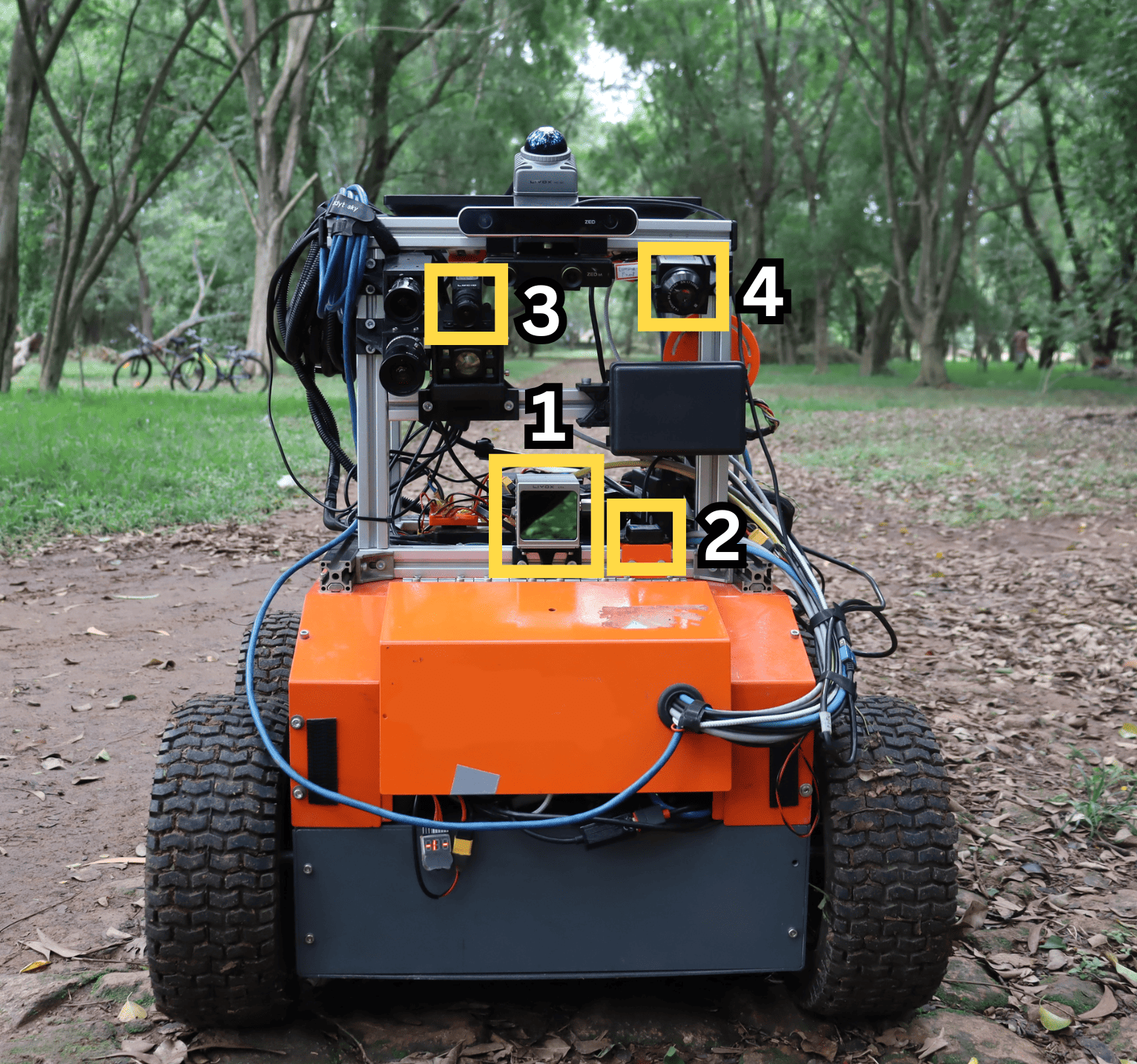}
        \captionsetup{font=footnotesize}
        \subcaption{Hardware Setup}
        \label{fig:hardware_setup}
    \end{minipage}\hfill
    \begin{minipage}[c]{0.63\textwidth}
        \centering
        
        \includegraphics[height=2.2cm, width=0.32\linewidth]{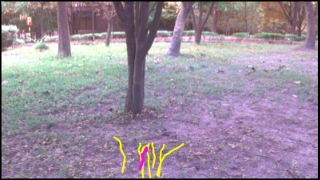}\hfill
        \includegraphics[height=2.2cm, width=0.32\linewidth]{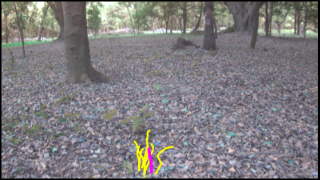}\hfill
        \includegraphics[height=2.2cm, width=0.32\linewidth]{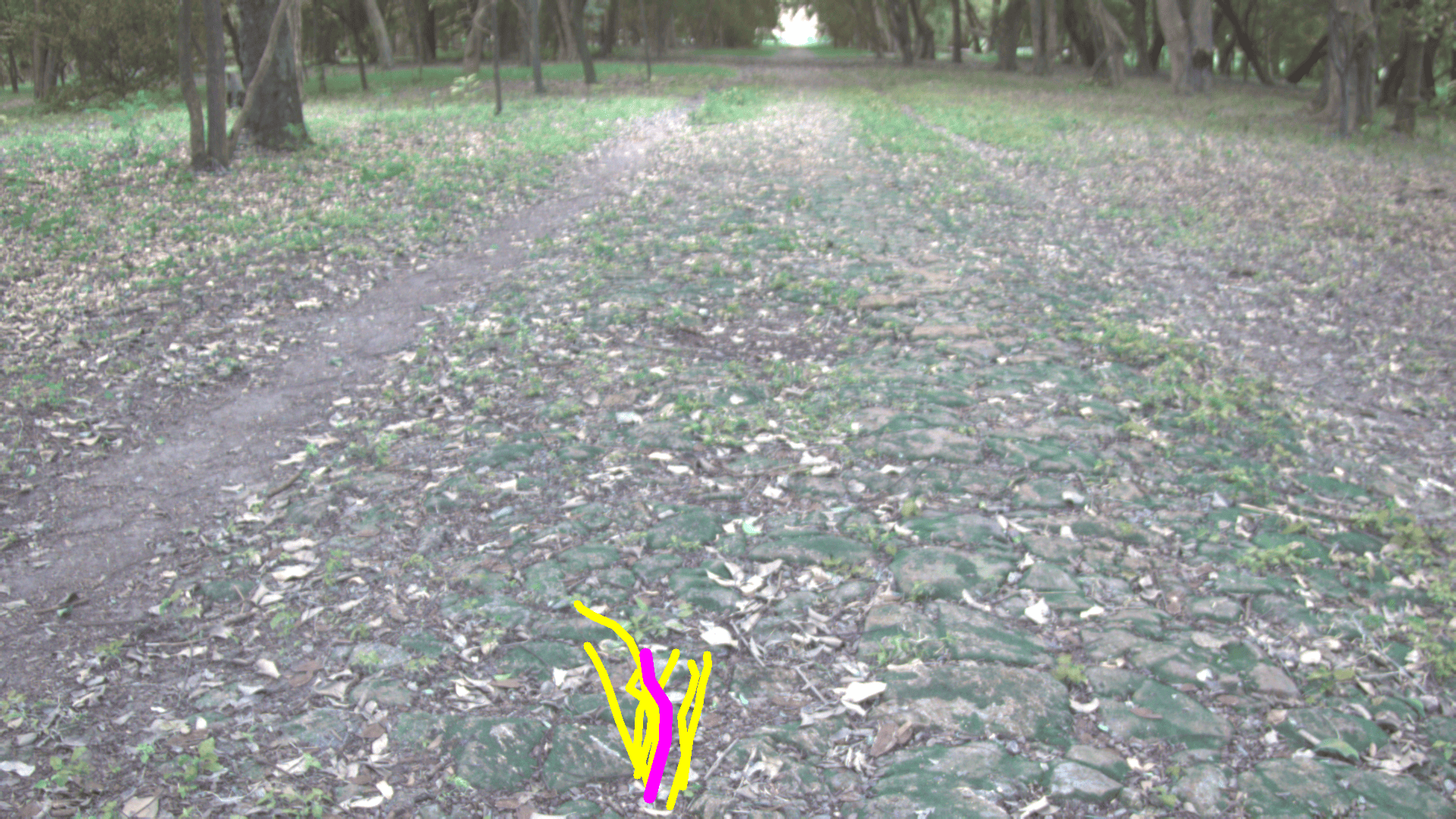}

        \vspace{0.2cm} 
        
                
        \includegraphics[height=2.2cm, width=0.32\linewidth]{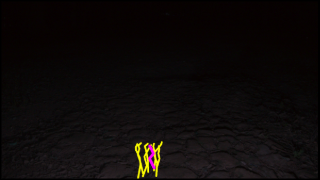}\hfill
        \includegraphics[height=2.2cm, width=0.32\linewidth]{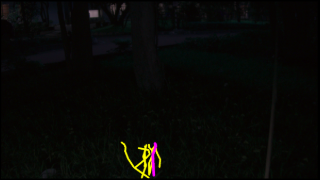}\hfill
        \includegraphics[height=2.2cm, width=0.32\linewidth]{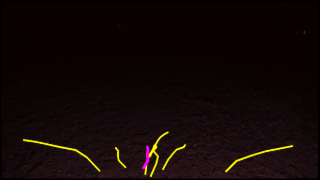}
        \captionsetup{font=footnotesize}
        \subcaption{Trajectory Rollouts in distinct off-road environments with varying illumination}
        \label{fig:traj_viz}
    \end{minipage}

    \captionsetup{font=footnotesize}
    \caption{\textbf{Hardware setup (left)} and \textbf{Visualizing trajectory rollouts (right).} 
    Figure (a) shows our hardware setup. Our sensor suite includes Livox Avia LiDAR, Wit IMU JY901, Lucid Vision Labs Triton TRI032S-CC Camera and Trimer TRA\_L6D\_ZDN LWIR Camera, indicated by 1,2,3 and 4 respectively. Figure (b) visualizes MAMMOTH's trajectory rollouts in (left) cluttered, (center) leafy with no clear trails, and (right) clear-trail environments. Top and bottom rows illustrate MAMMOTH's trajectory rollouts under daylight and low-ambient light conditions, respectively.}
    \vspace{-12pt}
    \label{fig:masking_viz}
\end{figure*}

We deployed all trained models on an all-terrain ground robot, Copernicus configured with our sensor suite as shown in \Cref{fig:hardware_setup}.
All onboard inference runs on a NVIDIA Jetson Orin AGX running ROS2 Humble. The forward pass runs at 2Hz. The robot has an average speed of 0.8m/s.

We deploy MAMMOTH across five distinct real-world off-road environments. We would like to highlight that the off-road environments chosen for testing and evaluation are significantly different from previous off-road navigation works like \cite{sivaprakasam2025salon}, \cite{10160856}, \cite{10740919} where robots are deployed in environments with distinctly clear trails, the environments are well-lit and there is a lack of overhead trees. As shown in our video, MAMMOTH was deployed during both day(adequate ambient light) and night(low ambient light) time in highly cluttered off-road, forest-like environments with dense overhead trees, no clear trails and complex terrain. \Cref{fig:traj_viz} shows MAMMOTH's trajectory rollouts in distinct environments in both day and night conditions.
We evaluate MAMMOTH by first comparing its performance to prior end-to-end methods in real-world visual-goal conditioned navigation and undirected exploration tasks. Subsequent analyses focus on the efficacy of the intrinsic traversability heuristic and the contribution provided by different sensor modality combinations to MAMMOTH's performance. We conclude our evaluation with a detailed ablation study of MAMMOTH’s internal architecture.

For all mobility experiments, we report the (i) mean collision rate (CR) which signifies mean number of collisions per minute of autonomous control (ii) goal-reaching success rate (SR), as taken from ~\cite{sridhar2024nomad} and (iii) mean number of “undesirable behaviors” (UB) which we define as the instance where the robot chooses a more unsafe and undesirable terrain in the distinct presence of a safer terrain and/or the robot operator had to take manual control, inspired by~\cite{sivaprakasam2025salon}. Hence, experiments for UB metric involved trail-following and reporting count of trail boundary violation instances along with manual takeover instances.

\subsection{Comparison with Baselines}
MAMMOTH is evaluated against two established baselines across five unseen off-road environments. These baselines are selected for their ability to address both visual-goal-conditioned navigation and undirected exploration tasks within a single policy. Because no prior end-to-end off-road methods satisfy the problem formulation and specifically address low-light conditions, standard RGB-only baselines are restricted to daytime deployment.
To ensure a fair, multi-modal comparison in both day and night settings, we adapt the best-performing RGB-only baseline to incorporate MAMMOTH's full sensor suite. We also introduce two ablation baselines, isolating the fusion technique and the intrinsic traversability heuristic, to validate our design decisions. All models share identical training datasets and splits. Performance is measured by: (i) reliable exploration during 40-minute worth autonomous mobility (excluding recovery time from collisions or manual takeovers) in day and night conditions each, and (ii) successful visual-goal navigation in daylight. Results are reported in \Cref{tab:e2e_comparison}.

\begin{enumerate}
    \item \textbf{NoMaD}: We train the authors implementation from scratch for 30 epochs and fine-tune for 5 epochs on our combined dataset. This is the most apt, easy-to-implement E2E learning-based baseline we found that closely matched the problem formulation of visual-goal navigation and undirected exploration. Furthermore, it was trained and shown to perform well in outdoor and off-road environments \cite{triest2022tartandrivelargescaledatasetlearning}. Both NoMaD and NoMaD-FT report inferior performance to MAMMOTH in all metrics proving that RGB-only methods are incapable of navigating complex off-road environments even in daytime conditions. Furthermore, they cannot generalize to low illumination where MAMMOTH reports even lower CR \& UB than NoMaD in daytime.
    
    \item \textbf{FlowNav}: This policy is built upon NoMaD and uses depth priors generated from Visual Foundational Models \cite{yang2024depth} as added modality along with conditional flow matching for trajectory generation. We train it from scratch for 30 epochs and fine-tune for 5 epochs. FlowNav and FlowNav-FT again prove inefficacy of RGB-only methods in complex off-road environments as adding depth priors derived from RGB images does not reduce CR as expected. 

    \item \textbf{Concat-fusion}: To validate the efficacy of the chosen fusion technique for MAMMOTH, we compare it against a "naive" baseline, where we switch the sparse MoE backbone for a simple concatenation operation and train with the same configuration as MAMMOTH. This method shows drastically poor performance for obstacle avoidance and terrain understanding (high CR and UB) in both day and night conditions, highlighting the importance of a strong fusion backbone.

    \item \textbf{Modified-NoMaD}: We adapt NoMaD to accept the same multi-modal inputs as MAMMOTH and inject the traversability heuristic into the diffusion head. The modality-specific encoders are identical to MAMMOTH. This was implemented as a fair multi-modal baseline and also to validate the efficacy of the chosen fusion technique compared to a dense transformer. While we observe a minor improvement in collision avoidance and terrain understanding from concat-fusion in both day and night time, CR \& UB values are still high which indicate that the vanilla dense transformer is insufficient for capturing the deep cross-modal understanding required for complex off-road mobility.

    \item \textbf{MAMMOTH w/o Heuristic}: To show the effectiveness of the intrinsic traversability heuristic injection into the diffusion policy, we compare MAMMOTH against a baseline with everything kept the same except the diffusion policy which models the distribution of geometric trajectories only (only coordinates of waypoints and no traversability heuristic) and train with the same configuration as MAMMOTH. While our multi-modal fusion enables MAMMOTH achieve decent obstacle avoidance indicated by relatively lower CR, high UB values indicate poor terrain understanding. After inclusion of the traversability heuristic into the diffusion policy, the model demonstrates better terrain awareness as indicated by substantial decrease in UB as well as decrease in CR in both day and night conditions. Furthermore, we would like to highlight that the intrinsic traversability heuristic injection has negligible computational overhead compared to external, post-hoc guidance. 
\end{enumerate}

As reported in \Cref{tab:e2e_comparison}, MAMMOTH outperforms all baselines and shows improvements in obstacle avoidance, better terrain understanding and daytime goal-reaching capability. Specifically, MAMMOTH surpasses the best-performing baseline, NoMaD \cite{sridhar2024nomad} by demonstrating a 0.6 reduction in CR and 16\% increase in SR during daytime. However, MAMMOTH's most impressive capability is its nighttime mobility since it not only remains operational when RGB-only baselines do not, but also significantly outperforms even the daytime metrics of all other baselines along with substantial reduction in the number of "undesirable behaviors". These promising results show MAMMOTH's superior capability in off-road mobility while underlining the importance of multi-modal perception. \Cref{fig:comparison} shows a failure case of NoMaD and FlowNav where MAMMOTH succeeds. We hypothesize that the overall low goal-reaching SR is due to the lack of distinct features in off-road environments, which reduces the quality of the temporal distance prediction.

\begin{table}[ht]
\caption{Comparison of E2E baselines on exploration and navigation tasks under day and night conditions. FT denotes that the policy is fine-tuned on our combined dataset.}
\label{tab:e2e_comparison}
\centering
\small
\renewcommand{\arraystretch}{1.1} 
\setlength{\tabcolsep}{2pt}    
\begin{tabular}{l ccccc} 
\specialrule{1pt}{0pt}{0pt} 
\multirow{3}{*}{\textbf{E2E Policy}} & \multicolumn{4}{c}{\textbf{Exploration}} & \multicolumn{1}{c}{\textbf{Navigation}} \\ 
\cmidrule(lr){2-5} \cmidrule(lr){6-6} 
& \multicolumn{2}{c}{\textbf{Day}} & \multicolumn{2}{c}{\textbf{Night}} & \multicolumn{1}{c}{\textbf{Day}} \\
\cmidrule(lr){2-3} \cmidrule(lr){4-5} \cmidrule(lr){6-6} 
& \textbf{CR $\downarrow$} & \textbf{UB $\downarrow$} & \textbf{CR$\downarrow$} & \textbf{UB$\downarrow$} & \textbf{SR(\%) $\uparrow$} \\ 
\specialrule{1pt}{0pt}{0pt} 
NoMaD \cite{sridhar2024nomad} & 1.3 & 1.0 & - & - & 17\% \\ 
NoMaD-FT \cite{sridhar2024nomad} & 2.1 & 1.7 & - & - & 20\% \\ 
FlowNav \cite{gode2025flownav}& 2.0 & 2.7 & - & - & 10\% \\ 
FlowNav-FT \cite{gode2025flownav} & 1.8 & 2.5 & - & - & 12\%\\ 
Concat-fusion & 1.7 & 1.6 & 1.7 & 2.1 & 14\% \\ 
Modified-NoMaD & 1.6 & 1.7 & 1.5 & 1.3 & 25\% \\ 
MAMMOTH w/o Heuristic  & 1.2 & 1.9 & 1.3 & 2.0 & 33\% \\ 
\rowcolor{cyan!10}
MAMMOTH & 0.7 & 0.6 & 0.8 & 0.75 & 33\% \\ 
\specialrule{1pt}{0pt}{0pt} 
\vspace{-24pt}
\end{tabular}
\end{table}

\subsection{Missing Modality Ablations}
To show the high deployability and robustness of MAMMOTH, we simulate missing modality conditions for obstacle avoidance and traversability estimation. These conditions were created by completely deactivating individual sensors to observe the impact of missing modalities for 12 minutes of continuous autonomous mobility for each experiment in day and night conditions each. Given the extensive range of possible sensor combinations, we evaluate two representative cases to characterize system performance: (i) single-modality dropout, to demonstrate MAMMOTH’s high deployability and robustness to individual sensor failures (ii) modality contribution analysis, by incrementally adding sensors to an RGB-only suite to evaluate the effectiveness of our fusion architecture.
As shown in \Cref{tab:masking1}, MAMMOTH performs best when none of the perception modalities (RGB, Thermal and 3D pointcloud) are missing while still generalizing well to one missing modality, particularly in night conditions. This shows that MAMMOTH builds strong cross-modal understanding capability. As expected, the 3D pointcloud modality plays a substantial role in collision avoidance, showing low CR when 3D pointcloud is present and high CR when 3D pointcloud is missing. We also observe the Thermal and Ego Velocity modalities to improve terrain understanding, especially at nighttime while 3D pointcloud shows no consistent role for learning traversability.

\Cref{tab:masking2} reports a sharp increase in CR and UB both, when RGB is the only perception modality active during both day and night conditions, again marking the importance of fusing other modalities. Addition of Thermal modality reduces UB while 3D pointcloud drastically reduces the CR in both day and night conditions.
This shows that our model “weighs” inputs from a combination of different modalities resulting in selective reliance according to environmental conditions to give good performance. Furthermore, MAMMOTH reports best performance when all modalities are available showing that it utilises information from different modalities in a complementary manner resulting in superior performance through synergistic fusion.

\begin{table}[t] 
\caption{Ablation study of single missing modalities for exploration tasks under day and night conditions.}
\label{tab:masking1}
\centering
\small 
\renewcommand{\arraystretch}{1.1} 
\setlength{\tabcolsep}{6pt}   
\begin{tabular}{l cccc} 
\specialrule{1pt}{0pt}{0pt} 
\multirow{2}{*}{\textbf{Missing Modalities}} & 
\multicolumn{2}{c}{\textbf{Day}} & 
\multicolumn{2}{c}{\textbf{Night}} \\ 
\cmidrule(lr){2-3} \cmidrule(lr){4-5}
& \textbf{CR} & \textbf{UB} & \textbf{CR} & \textbf{UB} \\ 
\specialrule{1pt}{0pt}{0pt} 
RGB & 1.1 & 1.0 & 0.8 & 1.5 \\ 
Thermal & 0.8 & 1.0 & 0.5 & 1.25 \\
3D pointcloud & 1.2 & 1.5 & 0.9 & 1.0 \\
Velocity & 0.75 & 0.75 & 0.9 & 1.0 \\
None & 0.7 & 0.6 & 0.8 & 0.75  \\
\specialrule{1pt}{0pt}{0pt} 
\vspace{-14pt}
\end{tabular}
\end{table}

\begin{table}[t] 
\caption{Ablation study of second modalities in addition to RGB for exploration tasks under day and night conditions.}
\label{tab:masking2}
\centering
\small 
\renewcommand{\arraystretch}{1.1} 
\setlength{\tabcolsep}{8pt}   
\begin{tabular}{l cccc} 
\specialrule{1pt}{0pt}{0pt} 
\multirow{2}{*}{\textbf{Available Modalities}} & 
\multicolumn{2}{c}{\textbf{Day}} & 
\multicolumn{2}{c}{\textbf{Night}} \\ 
\cmidrule(lr){2-3} \cmidrule(lr){4-5}
& \textbf{CR} & \textbf{UB} & \textbf{CR} & \textbf{UB} \\ 
\specialrule{1pt}{0pt}{0pt} 
RGB & 1.5 & 1.7 & 1.7 & 1.2\\
RGB + Thermal & 1.5 & 0.8 & 1.5 & 1.0 \\
RGB + 3D pointcloud & 0.7 & 1.0 & 0.7 & 1.7 \\
RGB + Velocity & 1.6 & 1.0 & 1.9 & 1.7 \\
All & 0.7 & 0.6 & 0.8 & 0.75  \\
\specialrule{1pt}{0pt}{0pt} 
\vspace{-24pt}
\end{tabular}
\end{table}

\begin{table}[ht]
\caption{Performance comparison of different architectural ablations.}
\label{tab::q4}
\centering
\small
\renewcommand{\arraystretch}{1.1} 
\setlength{\tabcolsep}{8pt}   
\begin{tabular}{l cccc} 
\specialrule{1pt}{0pt}{0pt} 
\multirow{2}{*}{\textbf{Ablation}} & 
\multicolumn{2}{c}{\textbf{Day}} & 
\multicolumn{2}{c}{\textbf{Night}} \\ 
\cmidrule(lr){2-3} \cmidrule(lr){4-5}
& \textbf{CR $\downarrow$} & \textbf{UB $\downarrow$} & \textbf{CR $\downarrow$} & \textbf{UB $\downarrow$} \\ 
\specialrule{1pt}{0pt}{0pt} 
Pretrained & 1.3 & 1.0 & - & - \\
Experts = 8 & 0.8 & 1.3 & 0.9 & 1.8 \\
Experts = 32 & 1.3 & 1.2 & 1.1 & 1.2 \\
Top-2 & 1.5 & 1.3 & 1.8 & 1.5 \\
Top-8 & 1.2 & 0.8 & 1.4 & 1.5 \\
Softmax gating & 1.0 & 0.8 & 1.2 & 0.75 \\
Joint router & 1.3 & 1.8 & 1.3 & 1.8 \\
\rowcolor{cyan!10}
Ours & 0.7 & 0.6 & 0.8 & 0.75 \\
\specialrule{1pt}{0pt}{0pt} 
\vspace{-24pt}
\end{tabular}
\end{table}

\subsection{Ablation Study}
We conduct the following ablations to validate MAMMOTH's architectural design decisions for 12 minutes
of continuous autonomous mobility for each experiment in day and night each. We refer to the configuration "Ours"  in \Cref{tab::q4} as 16 experts, $TopK = 4$, Laplace gating and per-modality router design.

{\bf Analysis of impact of pretrained vision encoder.}
We use a pretrained Efficientnet-B0 and compare it to our method where we trained the RGB and Thermal encoders end-to-end from scratch. We evaluate this method in day conditions only since it has no meaningful impact on nighttime evaluations. We report from \Cref{tab::q4}, clear inferior performance in both collision avoidance and terrain understanding. This may be attributed to the lack of off-road features in datasets EfficientNet-B0 was pretrained on \cite{tan2019efficientnet}.

{\bf Analysis of number of experts.}
We investigate the performance of MAMMOTH with different number of experts in the sparse MoE backbone. The default TopK value is 4. As shown in \Cref{tab::q4}, the use of 16 experts in our policy was the best choice of the three. While 8 experts shows marginally worse collision avoidance capability in both day and night conditions, it exhibits distinctly high UB, during both day and night conditions. We hypothesize that 8 experts may not be sufficient to capture the full representational complexity of all distinct modalities leading to unpredictable biases in its behavior. Moreover, we show similar results to  \cite{xu2023towards}, where increasing the number of experts in sparse MoE layers after a threshold may return diminishing results due to equal capacity assignment for the tokens.

{\bf Analysis of TopK values.}
We evaluate MAMMOTH’s performance with different TopK values in the sparse MoE backbone. The default number of experts is 16. \Cref{tab::q4} shows degraded performance when the TopK value is small ($TopK = 2$) due to inadequate contribution from various experts. A higher TopK value showed ($TopK = 8$) still showed inferior performance to $TopK = 4$, just as in \cite{guo2025dynamic}.

{\bf Analysis of gating functions and router design.}
We investigate the performance of MAMMOTH with different gating functions and router designs in the sparse MoE backbone. The number of experts is 16 and TopK value is 4 for all instantiations. \Cref{tab::q4} shows softmax gating as one the best performing ablations when it comes to terrain understanding, but the CR is still quite high. Moreover, \Cref{tab::q4} highlights the importance of modality specific routers compared to the classic single, joint router. This is consistent with \cite{han2024fusemoe} where laplace gating outperformed softmax and per-modality routers led to superior routing of tokens to experts for such multi-modal data.

\vspace{-8pt}
\section{Conclusion}
We propose MAMMOTH, a unified E2E policy for undirected exploration and visual-goal-conditioned navigation in off-road environments. Extensive experimental results demonstrate the strong performance of MAMMOTH in challenging off-road environments in varied illumination, highlighting the robustness,and deployability of the proposed policy. We further validate the efficacy and importance of our multi-modal fusion method and intrinsic traversability heuristic injection which helps to generate safer, reliable trajectories. We also demonstrate MAMMOTH's performance in various missing modality scenarios demonstrating flexibility and robustness to acute sensor degradation.

\bibliographystyle{IEEEtran}
\bibliography{root}

@String(CVPR  = {IEEE Conf. Comput. Vis. Pattern Recog.})

@String(CVPR  = {CVPR})

@inproceedings{sivaprakasam2025salon,
  title={Salon: Self-supervised adaptive learning for off-road navigation},
  author={Sivaprakasam, Matthew and Triest, Samuel and Ho, Cherie and Aich, Shubhra and Lew, Jeric and Adu, Isaiah and Wang, Wenshan and Scherer, Sebastian},
  booktitle={2025 IEEE International Conference on Robotics and Automation (ICRA)},
  pages={16999--17006},
  year={2025},
  organization={IEEE}
}

@inproceedings{10160856,
  author={Castro, Mateo Guaman and Triest, Samuel and Wang, Wenshan and Gregory, Jason M. and Sanchez, Felix and Rogers, John G. and Scherer, Sebastian},
  booktitle={2023 IEEE International Conference on Robotics and Automation (ICRA)}, 
  title={How Does It Feel? Self-Supervised Costmap Learning for Off-Road Vehicle Traversability}, 
  year={2023},
  volume={},
  number={},
  pages={931-938},
  doi={10.1109/ICRA48891.2023.10160856}
  }

@ARTICLE{10740919,
  author={Patel, Manthan and Frey, Jonas and Atha, Deegan and Spieler, Patrick and Hutter, Marco and Khattak, Shehryar},
  journal={IEEE Robotics and Automation Letters}, 
  title={RoadRunner M\&M - Learning Multi-Range Multi-Resolution Traversability Maps for Autonomous Off-Road Navigation}, 
  year={2024},
  volume={9},
  number={12},
  pages={11425-11432},
  doi={10.1109/LRA.2024.3490404}
  }

@misc{sharma2025irispathenhancingcostmapoffroad,
      title={IRisPath: Enhancing Costmap for Off-Road Navigation with Robust IR-RGB Fusion for Improved Day and Night Traversability}, 
      author={Saksham Sharma and Akshit Raizada and Suresh Sundaram},
      year={2025},
      eprint={2412.03173},
      archivePrefix={arXiv},
      primaryClass={cs.RO},
      url={https://arxiv.org/abs/2412.03173}, 
}

@inproceedings{shah2023vint,
  title={ViNT: A Foundation Model for Visual Navigation},
  author={Shah, Dhruv and Sridhar, Ajay and Dashora, Nitish and Stachowicz, Kyle and Black, Kevin and Hirose, Noriaki and Levine, Sergey},
  booktitle={Conference on Robot Learning},
  pages={711--733},
  year={2023},
  organization={PMLR}
}

@inproceedings{sridhar2024nomad,
  title={Nomad: Goal masked diffusion policies for navigation and exploration},
  author={Sridhar, Ajay and Shah, Dhruv and Glossop, Catherine and Levine, Sergey},
  booktitle={2024 IEEE International Conference on Robotics and Automation (ICRA)},
  pages={63--70},
  year={2024},
  organization={IEEE}
}

@INPROCEEDINGS{10802055,
  author={Liang, Jing and Payandeh, Amirreza and Song, Daeun and Xiao, Xuesu and Manocha, Dinesh},
  booktitle={2024 IEEE/RSJ International Conference on Intelligent Robots and Systems (IROS)}, 
  title={DTG : Diffusion-based Trajectory Generation for Mapless Global Navigation}, 
  year={2024},
  volume={},
  number={},
  pages={5340-5347},
  doi={10.1109/IROS58592.2024.10802055}
  }

@inproceedings{shazeer2017outrageously,
  title={Outrageously Large Neural Networks: The Sparsely-Gated Mixture-of-Experts Layer},
  author={Shazeer, Noam and Mirhoseini, Azalia and Maziarz, Krzysztof and Davis, Andy and Le, Quoc and Hinton, Geoffrey and Dean, Jeff},
  booktitle={International Conference on Learning Representations},
  year={2017}
}

@inproceedings{han2024fusemoe,
  title={FuseMoE: mixture-of-experts transformers for fleximodal fusion},
  author={Han, Xing and Nguyen, Huy and Harris, Carl and Ho, Nhat and Saria, Suchi},
  booktitle={Proceedings of the 38th International Conference on Neural Information Processing Systems},
  pages={67850--67900},
  year={2024}
}

@inproceedings{nezakati2025mmp,
  title={Mmp: Towards robust multi-modal learning with masked modality projection},
  author={Nezakati, Niki and Reza, Md Kaykobad and Patil, Ameya and Solh, Mashhour and Asif, M Salman},
  booktitle={2025 IEEE International Conference on Big Data (BigData)},
  pages={1480--1485},
  year={2025},
  organization={IEEE}
}

@inproceedings{shah2023gnm,
  title={Gnm: A general navigation model to drive any robot},
  author={Shah, Dhruv and Sridhar, Ajay and Bhorkar, Arjun and Hirose, Noriaki and Levine, Sergey},
  booktitle={2023 IEEE International Conference on Robotics and Automation (ICRA)},
  pages={7226--7233},
  year={2023},
  organization={IEEE}
}

@inproceedings{Shah_2022, series={RSS2022},
   title={ViKiNG: Vision-Based Kilometer-Scale Navigation with Geographic Hints},
   url={http://dx.doi.org/10.15607/RSS.2022.XVIII.019},
   DOI={10.15607/rss.2022.xviii.019},
   booktitle={Robotics: Science and Systems XVIII},
   publisher={Robotics: Science and Systems Foundation},
   author={Shah, Dhruv and Levine, Sergey},
   year={2022},
   month=jun, collection={RSS2022} }

@misc{yang2024pushinglimitscrossembodimentlearning,
      title={Pushing the Limits of Cross-Embodiment Learning for Manipulation and Navigation}, 
      author={Jonathan Yang and Catherine Glossop and Arjun Bhorkar and Dhruv Shah and Quan Vuong and Chelsea Finn and Dorsa Sadigh and Sergey Levine},
      year={2024},
      eprint={2402.19432},
      archivePrefix={arXiv},
      primaryClass={cs.RO},
      url={https://arxiv.org/abs/2402.19432}, 
}

@inproceedings{eftekharone,
  title={The One RING: a Robotic Indoor Navigation Generalist},
  author={Eftekhar, Ainaz and Weihs, Luca and Hendrix, Rose and Caglar, Ege and Salvador, Jordi and Herrasti, Alvaro and Han, Winson and VanderBilt, Eli and Kembhavi, Aniruddha and Farhadi, Ali and others},
  booktitle={The first CVPR workshop on 3D Vision Language Models (VLMs) for Robotics Manipulation: Opportunities and Challenges}
}

@inproceedings{tan2019efficientnet,
  title={Efficientnet: Rethinking model scaling for convolutional neural networks},
  author={Tan, Mingxing and Le, Quoc},
  booktitle={International conference on machine learning},
  pages={6105--6114},
  year={2019},
  organization={PMLR}
}

@inproceedings{lang2019pointpillars,
  title={Pointpillars: Fast encoders for object detection from point clouds},
  author={Lang, Alex H and Vora, Sourabh and Caesar, Holger and Zhou, Lubing and Yang, Jiong and Beijbom, Oscar},
  booktitle={Proceedings of the IEEE/CVF conference on computer vision and pattern recognition},
  pages={12697--12705},
  year={2019}
}

@article{ho2020denoising,
  title={Denoising diffusion probabilistic models},
  author={Ho, Jonathan and Jain, Ajay and Abbeel, Pieter},
  journal={Advances in neural information processing systems},
  volume={33},
  pages={6840--6851},
  year={2020}
}

@inproceedings{chi2023diffusionpolicy,
	title={Diffusion Policy: Visuomotor Policy Learning via Action Diffusion},
	author={Chi, Cheng and Feng, Siyuan and Du, Yilun and Xu, Zhenjia and Cousineau, Eric and Burchfiel, Benjamin and Song, Shuran},
	booktitle={Proceedings of Robotics: Science and Systems (RSS)},
	year={2023}
}

@misc{triest2022tartandrivelargescaledatasetlearning,
      title={TartanDrive: A Large-Scale Dataset for Learning Off-Road Dynamics Models}, 
      author={Samuel Triest and Matthew Sivaprakasam and Sean J. Wang and Wenshan Wang and Aaron M. Johnson and Sebastian Scherer},
      year={2022},
      eprint={2205.01791},
      archivePrefix={arXiv},
      primaryClass={cs.RO},
      url={https://arxiv.org/abs/2205.01791}, 
}

@inproceedings{sivaprakasam2024tartandrive,
  title={Tartandrive 2.0: More modalities and better infrastructure to further self-supervised learning research in off-road driving tasks},
  author={Sivaprakasam, Matthew and Maheshwari, Parv and Castro, Mateo Guaman and Triest, Samuel and Nye, Micah and Willits, Steve and Saba, Andrew and Wang, Wenshan and Scherer, Sebastian},
  booktitle={2024 IEEE International Conference on Robotics and Automation (ICRA)},
  pages={12606--12606},
  year={2024},
  organization={IEEE}
}

@inproceedings{shah2022rapid,
  title={Rapid Exploration for Open-World Navigation with Latent Goal Models},
  author={Shah, Dhruv and Eysenbach, Benjamin and Rhinehart, Nicholas and Levine, Sergey},
  booktitle={Conference on Robot Learning},
  pages={674--684},
  year={2022},
  organization={PMLR}
}

@inproceedings{gode2025flownav,
  title={Flownav: Combining flow matching and depth priors for efficient navigation},
  author={Gode, Samiran and Nayak, Abhijeet and Oliveira, D{\'e}bora NP and Krawez, Michael and Schmid, Cordelia and Burgard, Wolfram},
  booktitle={2025 IEEE/RSJ International Conference on Intelligent Robots and Systems (IROS)},
  pages={17762--17768},
  year={2025},
  organization={IEEE}
}

@misc{hirose2025omnivlaomnimodalvisionlanguageactionmodel,
      title={OmniVLA: An Omni-Modal Vision-Language-Action Model for Robot Navigation}, 
      author={Noriaki Hirose and Catherine Glossop and Dhruv Shah and Sergey Levine},
      year={2025},
      eprint={2509.19480},
      archivePrefix={arXiv},
      primaryClass={cs.RO},
      url={https://arxiv.org/abs/2509.19480}, 
}

@inproceedings{shah2023lm,
  title={Lm-nav: Robotic navigation with large pre-trained models of language, vision, and action},
  author={Shah, Dhruv and Osi{\'n}ski, B{\l}a{\.z}ej and Levine, Sergey and others},
  booktitle={Conference on robot learning},
  pages={492--504},
  year={2023},
  organization={pmlr}
}

@inproceedings{hiroselelan,
  title={LeLaN: Learning A Language-Conditioned Navigation Policy from In-the-Wild Video},
  author={Hirose, Noriaki and Glossop, Catherine and Sridhar, Ajay and Mees, Oier and Levine, Sergey},
  booktitle={8th Annual Conference on Robot Learning}
}

@misc{hirose2025learningdrivemodelbasedreannotation,
      title={Learning to Drive Anywhere with Model-Based Reannotation}, 
      author={Noriaki Hirose and Lydia Ignatova and Kyle Stachowicz and Catherine Glossop and Sergey Levine and Dhruv Shah},
      year={2025},
      eprint={2505.05592},
      archivePrefix={arXiv},
      primaryClass={cs.RO},
      url={https://arxiv.org/abs/2505.05592}, 
}

@inproceedings{liang2024mtg,
  title={Mtg: Mapless trajectory generator with traversability coverage for outdoor navigation},
  author={Liang, Jing and Gao, Peng and Xiao, Xuesu and Sathyamoorthy, Adarsh Jagan and Elnoor, Mohamed and Lin, Ming C and Manocha, Dinesh},
  booktitle={2024 IEEE International Conference on Robotics and Automation (ICRA)},
  pages={2396--2402},
  year={2024},
  organization={IEEE}
}

@inproceedings{thakker2025risk,
  title={Risk-Guided Diffusion: Toward Deploying Robot Foundation Models In Space, Where Failure Is Not An Option},
  author={Thakker, Rohan and Patnaik, Adarsh and Kurtz, Vince and Frey, Jonas and Becktor, Jonathan and Moon, Sangwoo and Royce, Rob and Kaufmann, Marcel and Georgakis, Georgios and Roth, Pascal and others},
  booktitle={RSS 2025 Workshop on Reliable Robotics: Safety and Security in the Face of Generative AI}
}

@inproceedings{zeng2025navidiffusor,
  title={Navidiffusor: Cost-guided diffusion model for visual navigation},
  author={Zeng, Yiming and Ren, Hao and Wang, Shuhang and Huang, Junlong and Cheng, Hui},
  booktitle={2025 IEEE International Conference on Robotics and Automation (ICRA)},
  pages={11994--12001},
  year={2025},
  organization={IEEE}
}

@inproceedings{yang2024diffusion,
  title={Diffusion-es: Gradient-free planning with diffusion for autonomous and instruction-guided driving},
  author={Yang, Brian and Su, Huangyuan and Gkanatsios, Nikolaos and Ke, Tsung-Wei and Jain, Ayush and Schneider, Jeff and Fragkiadaki, Katerina},
  booktitle={Proceedings of the IEEE/CVF conference on computer vision and pattern recognition},
  pages={15342--15353},
  year={2024}
}

@misc{liang2025mosuautonomouslongrangerobot,
      title={MOSU: Autonomous Long-range Robot Navigation with Multi-modal Scene Understanding}, 
      author={Jing Liang and Kasun Weerakoon and Daeun Song and Senthurbavan Kirubaharan and Xuesu Xiao and Dinesh Manocha},
      year={2025},
      eprint={2507.04686},
      archivePrefix={arXiv},
      primaryClass={cs.RO},
      url={https://arxiv.org/abs/2507.04686}, 
}

@inproceedings{nazeri2025verticoder,
  title={Verticoder: Self-supervised kinodynamic representation learning on vertically challenging terrain},
  author={Nazeri, Mohammad and Datar, Aniket and Pokhrel, Anuj and Pan, Chenhui and Warnell, Garrett and Xiao, Xuesu},
  booktitle={2025 IEEE International Conference on Robotics and Automation (ICRA)},
  pages={6536--6543},
  year={2025},
  organization={IEEE}
}

@inproceedings{datar2024terrain,
  title={Terrain-attentive learning for efficient 6-dof kinodynamic modeling on vertically challenging terrain},
  author={Datar, Aniket and Pan, Chenhui and Nazeri, Mohammad and Pokhrel, Anuj and Xiao, Xuesu},
  booktitle={2024 IEEE/RSJ International Conference on Intelligent Robots and Systems (IROS)},
  pages={5438--5443},
  year={2024},
  organization={IEEE}
}

@inproceedings{xu2023towards,
  title={Towards being parameter-efficient: a stratified sparsely activated transformer with dynamic capacity},
  author={Xu, Haoran and Elbayad, Maha and Murray, Kenton and Maillard, Jean and Goswami, Vedanuj},
  booktitle={Findings of the Association for Computational Linguistics: EMNLP 2023},
  pages={12858--12870},
  year={2023}
}

@inproceedings{guo2025dynamic,
  title={Dynamic mixture of experts: An auto-tuning approach for efficient transformer models},
  author={Guo, Yongxin and Cheng, Zhenglin and Tang, Xiaoying and Tu, Zhaopeng and Lin, Tao},
  booktitle={International Conference on Learning Representations},
  volume={2025},
  pages={79643--79672},
  year={2025}
}

@article{chen2021interpretable,
  title={Interpretable end-to-end urban autonomous driving with latent deep reinforcement learning},
  author={Chen, Jianyu and Li, Shengbo Eben and Tomizuka, Masayoshi},
  journal={IEEE Transactions on Intelligent Transportation Systems},
  volume={23},
  number={6},
  pages={5068--5078},
  year={2021},
  publisher={IEEE}
}

@misc{park2025endtoenddrivingselfsupervisedimitation,
      title={End-to-End Driving via Self-Supervised Imitation Learning Using Camera and LiDAR Data}, 
      author={Jin Bok Park and Jinkyu Lee and Muhyun Back and Hyunmin Han and David T. Ma and Sang Min Won and Sung Soo Hwang and Il Yong Chun},
      year={2025},
      eprint={2308.14329},
      archivePrefix={arXiv},
      primaryClass={cs.RO},
      url={https://arxiv.org/abs/2308.14329}, 
}

@INPROCEEDINGS{10658504,
  author={Lai, Lei and Ohn-Bar, Eshed and Arora, Sanjay and Yi, John Seon Keun},
  booktitle={2024 IEEE/CVF Conference on Computer Vision and Pattern Recognition (CVPR)}, 
  title={Uncertainty-Guided Never-Ending Learning to Drive}, 
  year={2024},
  volume={},
  number={},
  pages={15088-15098},
  doi={10.1109/CVPR52733.2024.01429}}

@article{yang2024depth,
  title={Depth anything v2},
  author={Yang, Lihe and Kang, Bingyi and Huang, Zilong and Zhao, Zhen and Xu, Xiaogang and Feng, Jiashi and Zhao, Hengshuang},
  journal={Advances in Neural Information Processing Systems},
  volume={37},
  pages={21875--21911},
  year={2024}
}

\end{document}